\documentclass[10pt,a4paper]{article}
\usepackage[a4paper,textwidth=12.2cm,textheight=19.3cm,centering]{geometry}
\usepackage{mathptmx}
\usepackage[T1]{fontenc}
\usepackage{amsmath,amssymb,graphicx,booktabs,algorithm,algpseudocode,xcolor,caption,subcaption,tikz}
\usetikzlibrary{arrows.meta,positioning,fit,calc}
\usepackage[hidelinks]{hyperref}
\newcommand{\method}{LiLib}
\title{\Large\bfseries \method: Lifelong Air-to-Ground Path-Loss Prediction\\on UAVs via a Drift-Triggered Model Library}
\author{Minh Tran\\[2pt]
\small Hanoi University of Science and Technology (HUST), Hanoi, Vietnam}
\date{}

\begin{document}
\maketitle

\begin{abstract}
UAVs that act as relays or base stations need accurate air-to-ground path-loss predictions for rate adaptation and placement, but propagation conditions change as a UAV moves between suburban, urban and high-rise areas, and the same areas are often revisited. Online regressors that adapt by forgetting must relearn each environment from scratch, whereas a single model trained on all data averages incompatible regimes. We propose \method{}, a lightweight continual-learning scheme in which a UAV maintains a small library of recursive-least-squares experts. A windowed residual test detects drift; a short probe phase then either reuses the best stored expert or creates a new one. In simulations based on four standard urbanization profiles, \method{} reduces prediction RMSE from 5.89 dB (best sliding-window baseline) to 4.03 dB ($p<0.001$), lowers the error shortly after a return to a known environment from 12.3 dB to 5.7 dB, and recovers 99\% of the throughput of a regime-aware oracle in rate adaptation. The library stores four experts in under 0.5\,KB, and identifies regimes with 92\% purity without labels. When a second UAV is initialized with the library of a peer, its error after environment changes halves. \method{} does not reach the oracle, and similar regimes may be merged when shadowing is strong. The results indicate that, for recurring drift, remembering is more effective than re-adapting.
\\[4pt]
\end{abstract}

\section{Introduction}
Unmanned aerial vehicles (UAVs) are increasingly used as aerial base stations, relays,
and connected users \cite{zeng2016,mozaffari2019}. Many of their networking functions,
such as link rate adaptation, handover, and placement, rely on predictions of the
air-to-ground (A2G) path loss. A2G propagation, however, depends strongly on the
surrounding environment: the probability of a line-of-sight (LoS) link and the excess
loss of non-LoS links differ markedly between suburban, urban, and high-rise areas
\cite{alhourani2014,3gpp36777}. A UAV performing a patrol, delivery, or monitoring
mission therefore experiences a \emph{non-stationary} data stream whose distribution
changes abruptly when it crosses into a different area and typically \emph{recurs}
when it returns.

Data-driven path-loss models can outperform empirical formulas when sufficient data
from the target environment are available \cite{seretis2022,phillips2013}, but they are
usually trained offline for a fixed environment. On a UAV, the model must instead be
learned and updated online, under tight computation and memory budgets. Two simple
strategies are common in practice. An adaptive regressor with a forgetting mechanism,
such as a sliding window or exponentially weighted least squares \cite{haykin2014},
tracks the current environment but discards previous knowledge, so every return to a
known area incurs the full relearning cost. Conversely, a single model fitted to all
past data does not forget, but because the environment is not observed directly, it
averages incompatible regimes. This is the stability--plasticity dilemma studied in
continual learning \cite{parisi2019} and concept-drift adaptation \cite{gama2014}.

This paper investigates whether a minimal continual-learning mechanism can resolve this
dilemma for A2G path-loss prediction on board a UAV. We propose \method{}, which
maintains a small \emph{library} of linear experts trained by recursive least squares
(RLS). A windowed test on the prediction residual signals drift; a short probe phase
then compares all stored experts on fresh samples and either \emph{reuses} the best
one or \emph{spawns} a new expert. The idea is related to mixtures of experts
\cite{jacobs1991} and expert-based lifelong learning \cite{aljundi2017}, but is
deliberately reduced to a form that runs with a few hundred bytes of memory and
requires no environment labels.

The contributions of this paper are as follows.
\begin{itemize}
\item We formulate on-board A2G path-loss prediction under recurring environment
changes as a task-free continual-learning problem and propose \method, a label-free
drift-triggered expert library with constant per-sample cost.
\item We show through simulation based on four standard urbanization profiles that
\method{} reduces prediction error relative to forgetting-based and cumulative
baselines, mainly by eliminating relearning after returns to known environments, and
that this translates into higher throughput and lower outage in rate adaptation.
\item We show that libraries can be shared between UAVs at negligible communication
cost, and we analyse the sensitivity of \method{} to its window length and to the
shadowing level, including the conditions under which it fails to separate regimes.
\end{itemize}

\section{Related Work}
\textbf{A2G channel modelling.} Al-Hourani \emph{et al.} \cite{alhourani2014} model the
A2G path loss as a mixture of LoS and non-LoS components whose probability depends on
the elevation angle and on environment-specific parameters; 3GPP specifies
scenario-dependent models for aerial users \cite{3gpp36777}. These models assume that
the environment type is known. Machine-learning approaches to propagation modelling
are reviewed in \cite{seretis2022,phillips2013}; they are typically trained offline for
a given deployment. The same holds for recent data-driven models of received signal
strength, such as transformer-based Wi-Fi RSSI fingerprinting for indoor positioning
\cite{nguyen2026rssi}.

\textbf{Continual learning and concept drift.} Continual learning studies how to
acquire new knowledge without catastrophic forgetting \cite{parisi2019}, using
regularization \cite{kirkpatrick2017}, replay \cite{rolnick2019}, or architectural
expansion such as networks of experts \cite{aljundi2017}. Concept-drift research
provides detectors such as the Page--Hinkley test \cite{page1954} and adaptive windowing
\cite{bifet2007}, surveyed in \cite{gama2014}. In deployed perception systems, adaptive
auto-learning has been used to keep a camera-based recognizer up to date after
deployment \cite{nguyen2025auto}. Most of this work targets deep networks
or general data streams. In contrast, we exploit the structure of the A2G problem, in
which a low-dimensional linear-in-parameters model suffices within one environment,
so that a library of RLS experts provides forgetting-free continual learning at very
low cost. Regularization-based methods such as EWC \cite{kirkpatrick2017} are not
directly suitable here, because the environment is not observed and a single
parameter vector cannot represent several regimes simultaneously.

\section{System Model and Problem Formulation}
\begin{figure}[t]
\centering
\includegraphics[width=0.55\linewidth]{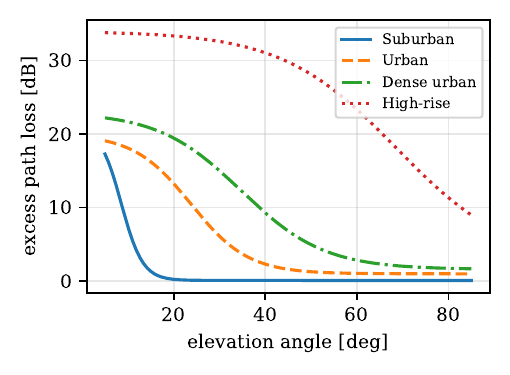}
\caption{Excess A2G path loss versus elevation angle for the four environments of
\cite{alhourani2014}. The environments form distinct regimes that a single model cannot
fit simultaneously.}\label{fig:envs}
\end{figure}

\textbf{Propagation model.} A UAV at altitude $h$ communicates with ground nodes at
horizontal distance $d$. Following \cite{alhourani2014}, the mean path loss in
environment $e$ is
\begin{gather}
\overline{\mathrm{PL}}_e(d,h)=20\log_{10}\frac{4\pi f_c r}{c}
+P^e_{\mathrm{LoS}}(\theta)\,\eta^e_{\mathrm{LoS}}+\bigl(1-P^e_{\mathrm{LoS}}(\theta)\bigr)\eta^e_{\mathrm{NLoS}},\\
P^e_{\mathrm{LoS}}(\theta)=\frac{1}{1+a_e e^{-b_e(\theta-a_e)}},\nonumber
\end{gather}
where $r=\sqrt{d^2+h^2}$, $\theta$ is the elevation angle in degrees, and $f_c=2$\,GHz.
We use the four environments of \cite{alhourani2014}: suburban
$(a,b,\eta_{\mathrm{LoS}},\eta_{\mathrm{NLoS}})=(4.88,0.43,0.1,21)$, urban
$(9.61,0.16,1,20)$, dense urban $(12.08,0.11,1.6,23)$, and high-rise urban
$(27.23,0.08,2.3,34)$. Fig.~\ref{fig:envs} shows that their excess losses differ by up
to about 20\,dB at low elevation angles. A measurement is
$y_t=\overline{\mathrm{PL}}_{e_t}(d_t,h_t)+\epsilon_t$ with log-normal shadowing
$\epsilon_t\sim\mathcal N(0,\sigma^2)$ and $\sigma=3$\,dB unless stated otherwise.

\textbf{Mission and data stream.} The UAV flies through a sequence of environment
segments, each yielding 300 measurements with $d\sim\mathcal U[20,1500]$\,m and
$h\sim\mathcal U[50,300]$\,m. A mission consists of three cycles, each visiting the four
environments in a random order, so that every environment recurs. The environment label
$e_t$ is \emph{not} available to the learner.

\textbf{Objective.} At each time $t$, the UAV predicts $\hat y_t$ from the features
$x_t$ before observing $y_t$, and then updates its model. The goal is to minimize the
prequential (test-then-train) root-mean-square error (RMSE) over the whole stream,
subject to a constant per-sample computation and a small, bounded memory.

\section{Proposed Method: \method}
\begin{figure}[t]
\centering
\begin{tikzpicture}[font=\scriptsize,
  box/.style={draw, rounded corners=2pt, align=center, minimum height=8mm, text width=24mm, fill=white},
  arr/.style={-{Stealth[length=2mm]}, thick}]
\node[box, fill=gray!10]   (in)    at (0,0)     {Measurement\\$x_t=(d,h,\theta)$, $y_t$};
\node[box, fill=green!12]  (exp)   at (3.6,0)   {Active expert $f_k$\\(RLS, 5 weights)};
\node[box]                 (out)   at (7.4,0)   {Prediction $\hat y_t$\\$\rightarrow$ rate adaptation};
\node[box, fill=red!10]    (det)   at (3.6,-1.7){Drift test\\$\frac{1}{W}\sum|e|>\tau$\,?};
\node[box, fill=yellow!20, text width=30mm] (probe) at (3.6,-3.3){Probe $W$ fresh samples:\\$j^*=\arg\min_j \mathrm{MAE}(f_j)$};
\node[box, fill=blue!8]    (reuse) at (0,-3.3)  {Reuse $f_{j^*}$\\(if MAE $<\alpha$)};
\node[box, fill=blue!8]    (spawn) at (7.4,-3.3){Spawn new expert\\(warm start: $f_{j^*}$)};
\node[box, fill=green!6, text width=86mm] (lib) at (3.7,-4.85) {Library $\mathcal L=\{f_1,\dots,f_M\}$ \quad ($\approx$120\,B per expert)};
\node[box, fill=violet!10] (peer)  at (3.7,-6.3) {Peer UAV library};
\draw[arr] (in) -- node[above]{predict} (exp);
\draw[arr] (exp) -- (out);
\draw[arr] (exp) -- node[left]{residual $e_t$} (det);
\draw[arr] (det) -- node[right]{alarm} (probe);
\draw[arr] (probe) -- (reuse);
\draw[arr] (probe) -- (spawn);
\draw[arr] (reuse) -- (reuse |- lib.north);
\draw[arr] (spawn) -- (spawn |- lib.north);
\draw[arr, dashed] (lib.east) -| (9.3,-0.85) -- node[above, pos=0.55]{set active expert} (4.5,-0.85) -- (4.5,-0.4);
\draw[thick, violet, {Stealth[length=2mm]}-{Stealth[length=2mm]}, dashed] (lib) -- node[right]{share ($\le$0.5\,KB)} (peer);
\end{tikzpicture}
\caption{Overview of \method. The active expert is updated by RLS at every step. When
the windowed residual exceeds $\tau$, a probe phase compares all experts on fresh
samples and either reuses the best one or spawns a new one. Libraries can be
shared with peer UAVs.}\label{fig:overview}
\end{figure}
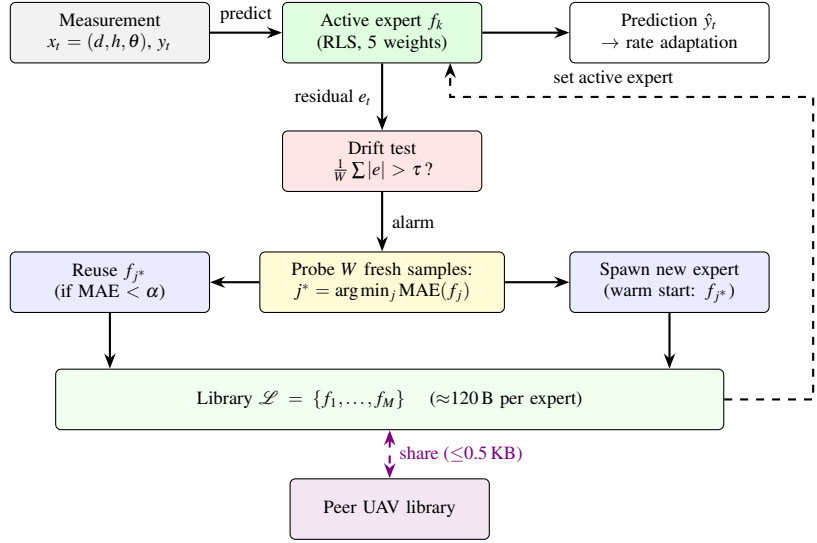

Fig.~\ref{fig:overview} summarizes \method. The UAV predicts with one \emph{active}
expert from a library $\mathcal L=\{f_1,\dots,f_M\}$, monitors its residual, and,
when drift is detected, decides between reusing a stored expert and creating a new one.

\subsection{Experts}
Each expert is a linear model $f_j(x)=w_j^\top\phi(x)$ with the five-dimensional basis
\begin{equation}
\phi(x)=\bigl[1,\ \log_{10}r-2.7,\ \tilde\theta,\ \tilde\theta^2,\ e^{-\theta/10}\bigr]^\top,
\qquad \tilde\theta=\theta/90-0.3,
\end{equation}
which captures the free-space term and a smooth approximation of the elevation-dependent
LoS mixture; the constants centre the features to improve numerical conditioning. Each
expert is trained by RLS without forgetting \cite{haykin2014}, which requires
$\mathcal O(D^2)$ operations per sample for $D=5$ and stores $w_j$ and a $5\times5$
covariance matrix (30 floats, 120\,B).

\subsection{Drift Detection, Probing, and Selection}
Only the active expert $f_k$ is updated. The UAV monitors the mean absolute residual
over the last $W$ samples and raises an alarm if it exceeds $\tau$, provided that
$f_k$ has seen at least $n_{\min}=60$ samples (to prevent alarms from immature experts).
After an alarm, \method{} enters a probe phase of $W$ samples. During this phase each
prediction is made by the expert with the lowest mean absolute error (MAE) on the probe
samples collected so far, and no expert is updated. At the end of the phase, let
$j^*=\arg\min_j \mathrm{MAE}_j$ on the probe samples. If $\mathrm{MAE}_{j^*}<\alpha$,
the UAV \emph{reuses} $f_{j^*}$; otherwise it \emph{spawns} a new expert initialized
with the weights of $f_{j^*}$ and a fresh covariance. In both cases the selected expert
is then updated with the probe samples. Algorithm~\ref{alg} summarizes the procedure.

The thresholds are expressed in terms of the shadowing standard deviation, which can
be taken from standard channel models: $\tau=1.5\sigma$ and $\alpha=1.17\sigma$ (4.5 and
3.5\,dB for $\sigma=3$\,dB). Since the MAE of a well-fitted expert is approximately
$0.8\sigma$, the alarm fires when the residual is about twice its nominal level. We
use $W=20$ unless stated otherwise.

\begin{algorithm}[t]
\caption{\method{} (per sample)}\label{alg}
\small
\begin{algorithmic}[1]
\If{in probe phase}
  \State predict with $\arg\min_j$ MAE$_j$ on probe buffer; append $(x_t,y_t)$
  \If{buffer full ($W$ samples)}
     \If{$\min_j \mathrm{MAE}_j<\alpha$} \State $k\gets j^*$ \Comment{reuse}
     \Else \State add $f_{\mathrm{new}}\gets f_{j^*}$ (fresh covariance); $k\gets$ new \Comment{spawn}
     \EndIf
     \State update $f_k$ with buffer; leave probe phase
  \EndIf
\Else
  \State predict $\hat y_t=f_k(x_t)$; update $f_k$ by RLS; push $|y_t-\hat y_t|$ to window
  \If{window mean $>\tau$ \textbf{and} $n_k\ge n_{\min}$} enter probe phase \EndIf
\EndIf
\end{algorithmic}
\end{algorithm}

\textbf{Cost and sharing.} Per sample, \method{} performs one RLS update; during the
short probe phase it additionally evaluates $M$ linear models on at most $W$ samples.
With four experts, the library occupies 480\,B, so transmitting it to a peer UAV is
negligible compared with typical control traffic. A UAV that receives a library starts
in the probe phase and thus selects the most suitable shared expert immediately.

\section{Experimental Setup}
All results are averaged over ten random seeds, each defining a mission of
$3\times4$ segments (3600 samples); the first 50 samples are excluded from error
statistics. We compare:
\begin{itemize}
\item \textbf{Static}: an expert trained on the first segment and then frozen;
\item \textbf{Cumulative}: a single RLS expert without forgetting, equivalent to
retraining on all past data (an idealized replay baseline \cite{rolnick2019});
\item \textbf{Sliding window (SW-100, SW-300)}: ridge regression refitted on the last
100 or 300 samples, a standard forgetting-based tracker \cite{gama2014};
\item \textbf{Oracle regime ID}: one RLS expert per true environment, selected with the
true label and warm-started from the previous expert at the first visit. It is not
deployable and serves as a reference.
\end{itemize}
All methods use the same features. For rate adaptation, the UAV transmits with 23\,dBm
over 1\,MHz (noise power $-107$\,dBm) and selects the spectral efficiency
$\min\{\log_2(1+\widehat{\mathrm{SNR}}/\gamma_m),6\}$ with a 2\,dB margin $\gamma_m$;
a transmission fails if this exceeds the capacity of the realized channel. Differences
in RMSE are tested with Welch's $t$-test over seeds.

\section{Results}
\subsection{Prediction Accuracy}
Table~\ref{tab:main} reports the prequential RMSE. \method{} achieves 4.03\,dB,
compared with 5.89\,dB for the best forgetting-based baseline (SW-100; $p<0.001$) and
10.45\,dB for the cumulative model. The cumulative model confirms that a single
parameter vector cannot represent the four regimes, and the longer window (SW-300) is
worse than the shorter one because it mixes regimes for longer after each change.
The oracle remains better (3.24\,dB; $p<0.001$), with the shadowing floor at
$\sigma=3$\,dB.

\begin{table}[t]
\centering\small\setlength{\tabcolsep}{4pt}
\caption{Prequential RMSE (dB, mean over ten seeds; $\pm$ standard deviation for the
whole stream). ``After switch'' denotes the first 50 samples of a segment.}\label{tab:main}
\begin{tabular}{lcccccc}
\toprule
 & & \multicolumn{2}{c}{Whole segment} & \multicolumn{2}{c}{After switch} & \\
\cmidrule(lr){3-4}\cmidrule(lr){5-6}
Method & All & First visit & Revisit & First visit & Revisit & Experts\\
\midrule
Static          & 16.80 $\pm$ 2.25 & 17.06 & 16.67 & 19.23 & 16.76 & 1\\
Cumulative      & 10.45 $\pm$ 0.09 & 10.60 & 10.38 & 14.38 & 11.04 & 1\\
SW-300          & 9.05 $\pm$ 0.49  & 7.93  & 9.48  & 13.49 & 14.49 & 1\\
SW-100          & 5.89 $\pm$ 0.28  & 5.34  & 6.11  & 11.45 & 12.29 & 1\\
\textbf{\method} & \textbf{4.03 $\pm$ 0.14} & \textbf{4.69} & \textbf{3.66} & \textbf{10.02} & \textbf{5.70} & 4.0\\
\midrule
Oracle regime ID & 3.24 $\pm$ 0.03 & 3.53 & 3.09 & 5.33 & 3.13 & 4\\
\bottomrule
\end{tabular}
\end{table}

The breakdown in Table~\ref{tab:main} shows where the gain originates. Immediately after
a return to a known environment, \method{} reduces the error from 12.29\,dB (SW-100) to
5.70\,dB, because it reuses the stored expert instead of relearning. After a switch to
a \emph{new} environment, its advantage is small (10.02 vs.\ 11.45\,dB), since a new
expert must still be learned; the oracle is better here only because it is told when
the switch occurs. On average \method{} created exactly 4.0 experts, i.e., one per
environment, and the active expert agreed with the true environment for 92\% of the
samples (cluster purity), although no labels were used.

\begin{figure}[t]
\centering
\includegraphics[width=\linewidth]{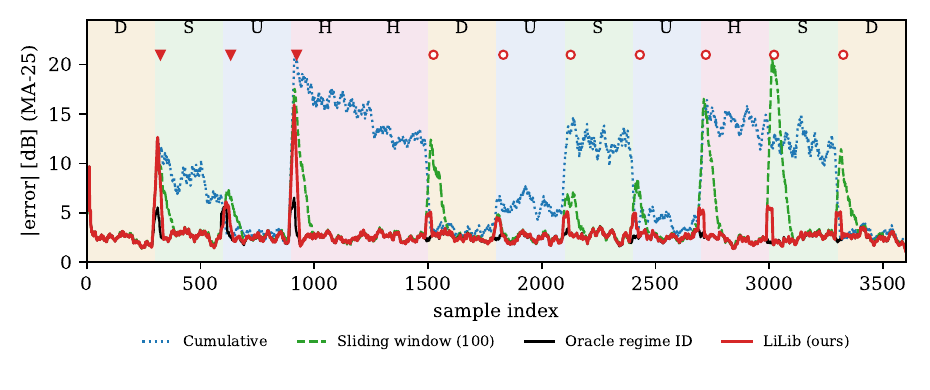}
\caption{Absolute prediction error (moving average over 25 samples) for one mission.
Background colours and letters indicate the true environment (S: suburban, U: urban,
D: dense urban, H: high-rise). Triangles: \method{} spawns an expert; circles:
\method{} reuses an expert.}\label{fig:stream}
\end{figure}

Fig.~\ref{fig:stream} illustrates one mission. The forgetting-based tracker exhibits an
error peak after every switch, including returns to known environments, while
\method{} shows short peaks corresponding to the detection and probe delay, after which
it reuses a stored expert (circles) or spawns a new one (triangles).

\subsection{Impact on Rate Adaptation}
Table~\ref{tab:rate} translates the prediction errors into link performance. \method{}
attains 3.36\,b/s/Hz with 17.9\% outage, compared with 3.23\,b/s/Hz and 21.1\% for
SW-100, and recovers 99\% of the oracle's throughput. The gap to the genie that knows
the realized SNR (4.21\,b/s/Hz) is dominated by shadowing, which no predictor based on
position alone can remove. The high outage of all methods reflects the fixed 2\,dB
margin, which is small relative to $\sigma=3$\,dB; a larger margin would trade
throughput for reliability but would not change the ranking.

\begin{table}[t]
\centering\small
\caption{Rate adaptation based on the predicted SNR (mean over ten seeds).}\label{tab:rate}
\begin{tabular}{lcc}
\toprule
Method & Throughput (b/s/Hz) & Outage (\%)\\
\midrule
Static           & 1.91 & 22.2\\
Cumulative       & 2.86 & 28.1\\
SW-300           & 2.91 & 28.3\\
SW-100           & 3.23 & 21.1\\
\textbf{\method} & \textbf{3.36} & \textbf{17.9}\\
\midrule
Oracle regime ID & 3.40 & 17.3\\
Genie (realized SNR) & 4.21 & 0.0\\
\bottomrule
\end{tabular}
\end{table}

\subsection{Sharing Libraries Between UAVs}
To assess knowledge transfer between UAVs, UAV A completes one cycle of the four
environments, and UAV B subsequently flies one cycle in a different order. Without
sharing, every environment is new to B. Table~\ref{tab:share} shows that initializing B
with A's library reduces its RMSE from 5.78 to 3.59\,dB and the error after
environment changes from 12.29 to 5.46\,dB. When A has visited only two environments,
B still benefits (4.17\,dB), reusing the two shared experts and spawning the two
missing ones. The transfer costs at most 480\,B.

\begin{table}[t]
\centering\small
\caption{UAV B's prediction error with and without A's library (one cycle; ten seeds).}\label{tab:share}
\begin{tabular}{lccc}
\toprule
Initialization of UAV B & RMSE (dB) & After switch (dB) & Data sent\\
\midrule
No sharing                       & 5.78 & 12.29 & 0\\
Library of A (2 environments)    & 4.17 & 7.45 & 240\,B\\
Library of A (4 environments)    & \textbf{3.59} & \textbf{5.46} & 480\,B\\
\bottomrule
\end{tabular}
\end{table}

\subsection{Sensitivity Analysis}
Table~\ref{tab:sens} varies the window length $W$ and the shadowing level $\sigma$ (with
thresholds scaled by $\sigma$). A shorter window ($W=10$) lowers the RMSE to 3.67\,dB by
reacting faster, but occasionally spawns redundant experts (4.4 on average); a longer
window ($W=40$) produces fewer experts but reacts more slowly (4.66\,dB). The default
$W=20$ balances the two. Under stronger shadowing ($\sigma=5$\,dB), \method{} creates
only 3.3 experts on average: the noise masks the difference between the urban and
dense-urban environments, whose excess losses are similar (Fig.~\ref{fig:envs}), and
they are often represented by a single expert. The gap to the oracle nevertheless
remains below 1\,dB in all cases.

\begin{table}[t]
\centering\small
\caption{Sensitivity of \method{} (RMSE in dB; number of experts in parentheses).}\label{tab:sens}
\begin{tabular}{lccc}
\toprule
Window length $W$ ($\sigma=3$\,dB) & 10 & 20 & 40\\
\method{} & 3.67 (4.4) & 4.03 (4.0) & 4.66 (3.7)\\
\midrule
Shadowing $\sigma$ ($W=20$) & 2\,dB & 3\,dB & 5\,dB\\
\method{} & 3.16 (4.1) & 4.03 (4.0) & 6.05 (3.3)\\
Oracle regime ID & 2.28 & 3.24 & 5.23\\
\bottomrule
\end{tabular}
\end{table}

\section{Discussion and Threats to Validity}
Several limitations qualify these findings. First, the data are synthetic and generated
by the same family of models from which the features were derived; real measurements
include terrain, antenna patterns, and spatially correlated shadowing, and the linear
experts may fit them less well. Second, environments change abruptly in our missions;
gradual transitions, such as a suburb blending into a city, may trigger alarms later or
produce intermediate experts. Third, the thresholds assume that $\sigma$ is known; a
misestimated $\sigma$ would lead to too many or too few experts, as the window
analysis suggests. Fourth, the library grows without bound in principle; although
it stayed at about four experts here, long deployments would require merging or pruning
rules. Finally, we did not compare with deep continual-learning methods, as our aim is
a method that fits within the memory of a microcontroller; the oracle results bound
the remaining improvement at under 1\,dB in our setting. Within these limits, the
results suggest that when drift is recurrent, the main benefit comes from remembering
previous regimes rather than from adapting faster.

\section{Conclusion}
This paper presented \method, a drift-triggered library of RLS experts for lifelong
A2G path-loss prediction on UAVs. Without environment labels, \method{} identifies
recurring propagation regimes, reuses stored experts after returning to known areas,
and reduces prediction error by about 1.9\,dB compared with the best forgetting-based
baseline, which improves rate adaptation. Its memory footprint is under 0.5\,KB, and
libraries can be shared between UAVs to accelerate learning. Future work will validate
the method on measured A2G data, handle gradual transitions, add library pruning, and
combine the predictor with UAV placement and trajectory design.

\end{document}